\documentclass[lettersize,journal]{IEEEtran}
\usepackage{amsmath,amsfonts}
\usepackage{algorithmic}
\usepackage{algorithm}
\usepackage{array}
\usepackage[caption=false,font=normalsize,labelfont=sf,textfont=sf]{subfig}
\usepackage{textcomp}
\usepackage{stfloats}
\usepackage{url}
\usepackage{verbatim}
\usepackage{graphicx}
\usepackage{cite}
\usepackage{tikz}
\usepackage{pgfplots}
\pgfplotsset{compat=newest}
\usepackage{booktabs}
\begin{document}

\title{EpiDiffVO: Geometry-Aware Epipolar Diffusion for Robust Visual Odometry}

\author{Prateeth Rao%
\thanks{
Prateeth Rao is with the International Institute of Information Technology Bangalore, Bengaluru, India 
(e-mail: prateeth.rao@iiitb.ac.in).
}
\thanks{Manuscript received April 19, 2021; revised August 16, 2021.}}

\markboth{Journal of \LaTeX\ Class Files,~Vol.~14, No.~8, August~2021}%
{Shell \MakeLowercase{\textit{et al.}}: A Sample Article Using IEEEtran.cls for IEEE Journals}


\maketitle

\begin{abstract}
Estimating relative pose from image pairs fundamentally requires only a minimal subset of geometrically consistent correspondences. However, most learning-based approaches rely on dense matching or direct regression, leading to redundancy and reduced geometric interpretability. In this work, we propose a sparse epipolar matching framework that predicts a compact set of correspondences optimized for geometric consistency across varying temporal baselines. To address residual noise and misalignment, we introduce an epipolar diffusion process that models correspondence uncertainty and refines keypoints toward epipolar consistency. The refined correspondences, along with depth cues, are lifted into a graph representation forming a Steiner graph that encodes relational structure between points. A graph neural network learns a compact subset of informative correspondences, which are passed to a differentiable singular value decomposition solver for end-to-end geometric estimation. Relative pose is recovered from the resulting essential matrix and evaluated in a visual odometry setting on the TartanAir and KITTI SLAM datasets. Experimental results demonstrate that combining sparse matching, diffusion-based refinement, and graph-based subset selection reduces correspondence redundancy while maintaining robust pose estimation across challenging baselines. Code and implementation details will be made publicly available at:
\url{https://github.com/Prateeth8/Epipolar-Diffusion-Pose}
\end{abstract}

\begin{IEEEkeywords}
Epipolar Geometry, Visual Odometry, Sparse Image Matching, DDPM, GNN, Sampson Loss, Steiner Graph.
\end{IEEEkeywords}

\section{Introduction}
\IEEEPARstart{T}{he} estimation of relative camera motion from image pairs is a foundational problem in visual odometry and robotics. Classical multi-view geometry shows that only a minimal set of correspondences is sufficient for pose recovery. However, modern learning-based approaches increasingly rely on dense and redundant matches, which introduces unnecessary complexity further being sensitive to noise. Lack of matched correspondences is not an issue in Visual Odometry, but identifying a geometrically consistent subset that reliably constrains motion under real-world conditions such as wide baselines, motion blur, and viewpoint changes.

A major source of failure in correspondence-based pipelines arises from deviations from ideal imaging assumptions. Real-world correspondences are affected by structured noise, including motion blur, illumination changes, calibration inaccuracies, and violations of strict epipolar alignment due to irregular spacing between images or dynamic objects in the scene. Existing methods typically treat these effects implicitly or assume simple noise models, limiting their ability to enforce geometric consistency under challenging conditions. As a result, even high-quality matches may not align with the underlying geometric constraints required for accurate pose estimation.

Recent learning-based approaches often attempt to directly regress motion parameters or essential matrices from correspondences. While effective in some settings, such formulations obscure the underlying geometric structure and can lead to physically inconsistent predictions. This issue is particularly pronounced in monocular visual odometry, where translation is inherently ambiguous up to scale. Direct regression entangles correspondence selection with motion estimation, making it difficult to disentangle geometric validity from data-driven biases and limiting generalization across diverse motion patterns.

Most of the image matching algorithms treat matches independently or rely on global aggregation mechanisms such as attention, without explicitly modeling the relational dependencies between correspondences. However, geometric consistency is inherently a structured property, where the validity of a match depends on its agreement with others. Ignoring this structure makes it difficult to identify minimal, informative subsets of correspondences required for robust pose estimation. These limitations suggest the need for a framework that (i) explicitly models structured uncertainty in correspondences, (ii) preserves geometric constraints without relying on direct regression, and (iii) leverages relational structure to identify minimal, informative subsets for pose estimation. In this work, we propose such a framework, combining probabilistic refinement, structured representation, and differentiable geometric estimation within a unified pipeline.

The proposed perspective is particularly relevant for robotic and aerial navigation scenarios, where large inter-frame motion, sparse texture, and environmental noise challenge conventional correspondence pipelines. We evaluate our approach on benchmarks such as TartanAir and TartanAir v2, demonstrating robustness under conditions representative of drone-based visual odometry.

\textbf{Contributions} of this paper are summarized as follows:
\begin{itemize}

\item We introduce a \textbf{sparse correspondence framework} that prioritizes geometrically consistent matches over dense predictions for robust pose estimation.

\item We propose a \textbf{diffusion-based keypoint realignment} strategy that models correspondence perturbations under an isotropic Gaussian process and refines matches to satisfy epipolar constraints.

\item We formulate correspondence aggregation as a \textbf{2D-3D Steiner graph construction}, leveraging stereo depth to capture both global structure (via MST) and local consistency (via KNN) using transformers with attention modules.

\item We design a \textbf{multi-hypothesis geometric estimation pipeline} that predicts multiple essential matrix candidates and recovers relative pose through a differentiable SVD-based solver.

\end{itemize}

\section{Related Work}

This section reviews prior work spanning the evolution from classical image matching to deep and diffusion-based approaches, examines the role and assumptions of classical solvers in pose estimation, and concludes with graph-based learning methods for improved correspondence and pose estimation.

\textbf{Image Matching :} Image matching traditionally relies on feature-based and area-based methods to establish correspondences between images ~\cite{imgmatchhandMa2021}. Feature-based approaches use keypoint detection, descriptor extraction, and nearest neighbour matching followed by geometric verification, while area-based methods directly compare image patches. Although feature-based methods are more robust to geometric variations, they suffer from combinatorial complexity and sensitivity to outliers. Area-based methods are computationally expensive and sensitive to illumination and noise ~\cite{imgmatchhandMa2021}. These limitations motivate the use of learned representations for improved robustness and scalability ~\cite{deepimgsurvey2024}.

\textbf{Deep Image Matching:} Deep learning methods learn feature representations and matching functions directly from data, improving robustness to viewpoint and appearance variations ~\cite{localfeatleng2019}. CNN-based approaches enable efficient feature extraction ~\cite{patch2pixzhou2021} ~\cite{xfeatpotje2024} ~\cite{sparsetimofte2015}, while graph-based methods such as SuperGlue~\cite{supersarlin2020} and LightGlue~\cite{lightlinden2023} model matching as a context-aware attention problem with optimal assignment. Transformer-based methods like COTR~\cite{cotrjiang2021} perform query-based correspondence prediction using global context. Dense matching approaches such as DKM~\cite{dkmEdstedt2023} further extend this to continuous correspondence estimation. Despite improved performance, these methods often lack explicit geometric constraints.

\textbf{Diffusion into Image Matching :} Recent works model correspondence estimation as a diffusion process, where matches are iteratively refined from noisy initial states~\cite{nam2024diffusion}. Classical non-linear diffusion methods perform edge-aware refinement of disparity ~\cite{stereononlinear1996}, while modern approaches learn to denoise correspondence fields using deep networks. Diffusion-based methods improve robustness to occlusions, ambiguities, and large viewpoint changes by enforcing consistency over multiple steps. Extensions such as DiffGlue~\cite{zhang2024diffglue} and keypoint-based diffusion ~\cite{6diffkey2024} frameworks further integrate refinement into feature matching and pose estimation pipelines.

\textbf{Classical Solvers :} Classical solvers estimate geometric models such as the fundamental or essential matrix from correspondences using robust optimization techniques. These methods are critical for enforcing geometric consistency and computing camera pose. However, their performance depends heavily on the quality of input matches and they rely on handcrafted heuristics for outlier rejection. Despite advances in learning-based methods, classical solvers remain integral for reliable geometric estimation ~\cite{ransacsurvey}.

\textbf{Graph Construction and Learning:} Graph-based approaches formulate feature matching as a relational problem, where keypoints are nodes and edges encode spatial or descriptor relationships. Graph neural networks propagate contextual information to improve matching consistency, as seen in methods like SuperGlue~\cite{supersarlin2020} and seeded graph matching~\cite{graphmatchchen2021}. These approaches reduce ambiguity in challenging regions by leveraging global structure. Hybrid methods further integrate graph reasoning with geometric estimation for iterative refinement of correspondences~\cite{stereogluebarath2024}.

\begin{figure*}
    \centering
    \includegraphics[width=0.75\linewidth]{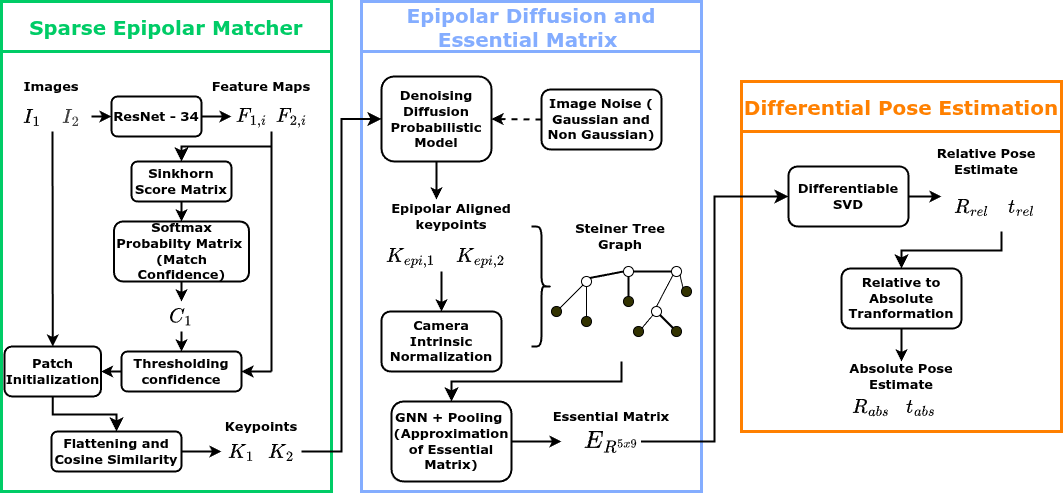}
    \caption{Geometry Diffusion VO module: a) Sparse Epipolar Matcher for estimating initial subset of matched keypoints, b) Epipolar Diffusion and Graph Construction: Keypoint Realignment process using DDPM with depth upliftment based Steiner Graph Construction and c) Differential SVD based Pose estimation}
    \label{fig:epi-block-pose}
\end{figure*}

\section{Preliminaries}

The Preliminaries section presents the foundational concepts underlying the proposed framework, including: a) detection-free image matching, b) noise affecting epipolar alignment, c) minimal solvers for pose estimation, and d) Steiner tree-based graph construction. These subsections establish the mathematical and theoretical basis for the proposed methodology.

\subsection{Feature Map based Image Matching}

Let $I_1, I_2 \in \mathbb{R}^{H \times W \times 3}$ denote a stereo image pair, where $H=480$ and $W=640$. These images are passed through a feature extraction network $M_1$ (ResNet-34) to obtain intermediate feature maps:
\begin{equation}
F_1 = M_1(I_1), \quad F_2 = M_1(I_2),
\end{equation}
where $F_1, F_2 \in \mathbb{R}^{\frac{H}{24} \times \frac{W}{32} \times 256}$.

\vspace{2mm}

\noindent \paragraph{Score Matrix Computation}  
Feature maps are reshaped into a sequence of feature vectors:
\begin{equation}
\hat{F}_1 \in \mathbb{R}^{N_1 \times d}, \quad \hat{F}_2 \in \mathbb{R}^{N_2 \times d},
\end{equation}
where $d=256$, $N_1=\frac{H}{24}\cdot\frac{W}{32}$ and $N_2=\frac{H}{24}\cdot\frac{W}{32}$.

The similarity score matrix is computed using dot-product similarity:
\begin{equation}
S_1 = \hat{F}_1 \hat{F}_2^\top,
\end{equation}
where $S_1 \in \mathbb{R}^{N_1 \times N_2}$.

\vspace{2mm}

\noindent \paragraph{Sinkhorn Normalization}  
The score matrix is converted into a probabilistic matching matrix using Sinkhorn normalization:
\begin{equation}
P_1 = \text{Sinkhorn}(S_1),
\end{equation}
where $P_1$ is a doubly stochastic matrix:
\begin{equation}
P_1 \mathbf{1} = \mathbf{1}, \quad P_1^\top \mathbf{1} = \mathbf{1}.
\end{equation}

\vspace{2mm}

\noindent \paragraph{Thresholding}  
A confidence threshold $\tau = 0.65$ is applied to retain reliable correspondences:
\begin{equation}
\mathcal{M} = \{(i,j) \mid P_1(i,j) > \tau \}.
\end{equation}

\vspace{2mm}

\noindent \paragraph{Positional Encoding}  
For each valid match $(i,j) \in \mathcal{M}$, positional encodings are added:
\begin{equation}
\tilde{F}_1(i) = \hat{F}_1(i) + \text{PE}(i), \quad
\tilde{F}_2(j) = \hat{F}_2(j) + \text{PE}(j),
\end{equation}
where $\text{PE}(\cdot)$ denotes positional encoding.

\vspace{2mm}

\noindent \paragraph{Patch Projection}  
Matched features are projected back to the image space as local patches:
\begin{equation}
\mathcal{P}_1 = \{ \phi(I_1, i) \}, \quad \mathcal{P}_2 = \{ \phi(I_2, j) \},
\end{equation}
where $\phi(\cdot)$ extracts a patch centered at the corresponding location.

Each patch is flattened into a vector:
\begin{equation}
p_1^k, p_2^k \in \mathbb{R}^{d_p}.
\end{equation}

\vspace{2mm}

\noindent \paragraph{Cosine Similarity Matching}  
Final matching scores are computed using cosine similarity:
\begin{equation}
\text{sim}(p_1^k, p_2^k) = \frac{p_1^k \cdot p_2^k}{\|p_1^k\| \|p_2^k\|}.
\end{equation}

Top-$K$ matches are selected:
\begin{equation}
\mathcal{M}_{\text{final}} = \text{TopK}(\text{sim}(p_1^k, p_2^k)).
\end{equation}

\subsection{Noise Affecting Epipolar Alignment}

Let $I_1$ and $I_2$ be a stereo-initialized image pair with corresponding keypoints $\mathbf{x}_1, \mathbf{x}_2 \in \mathbb{R}^2$. Under ideal conditions:
\begin{equation}
\mathbf{x}_2^\top \mathbf{E} \mathbf{x}_1 = 0
\end{equation}
where $\mathbf{E} = [\mathbf{t}]_\times \mathbf{R}$.

Observed correspondences are perturbed:
\begin{equation}
\tilde{\mathbf{x}}_i = \mathbf{x}_i + \boldsymbol{\epsilon}_i, \quad i \in \{1,2\}
\end{equation}

\begin{equation}
\boldsymbol{\epsilon} = 
\boldsymbol{\epsilon}^{proj} +
\boldsymbol{\epsilon}^{motion} +
\boldsymbol{\epsilon}^{match} +
\boldsymbol{\epsilon}^{calib}
\end{equation}

\begin{figure}
    \centering
    \includegraphics[width=0.85\linewidth]{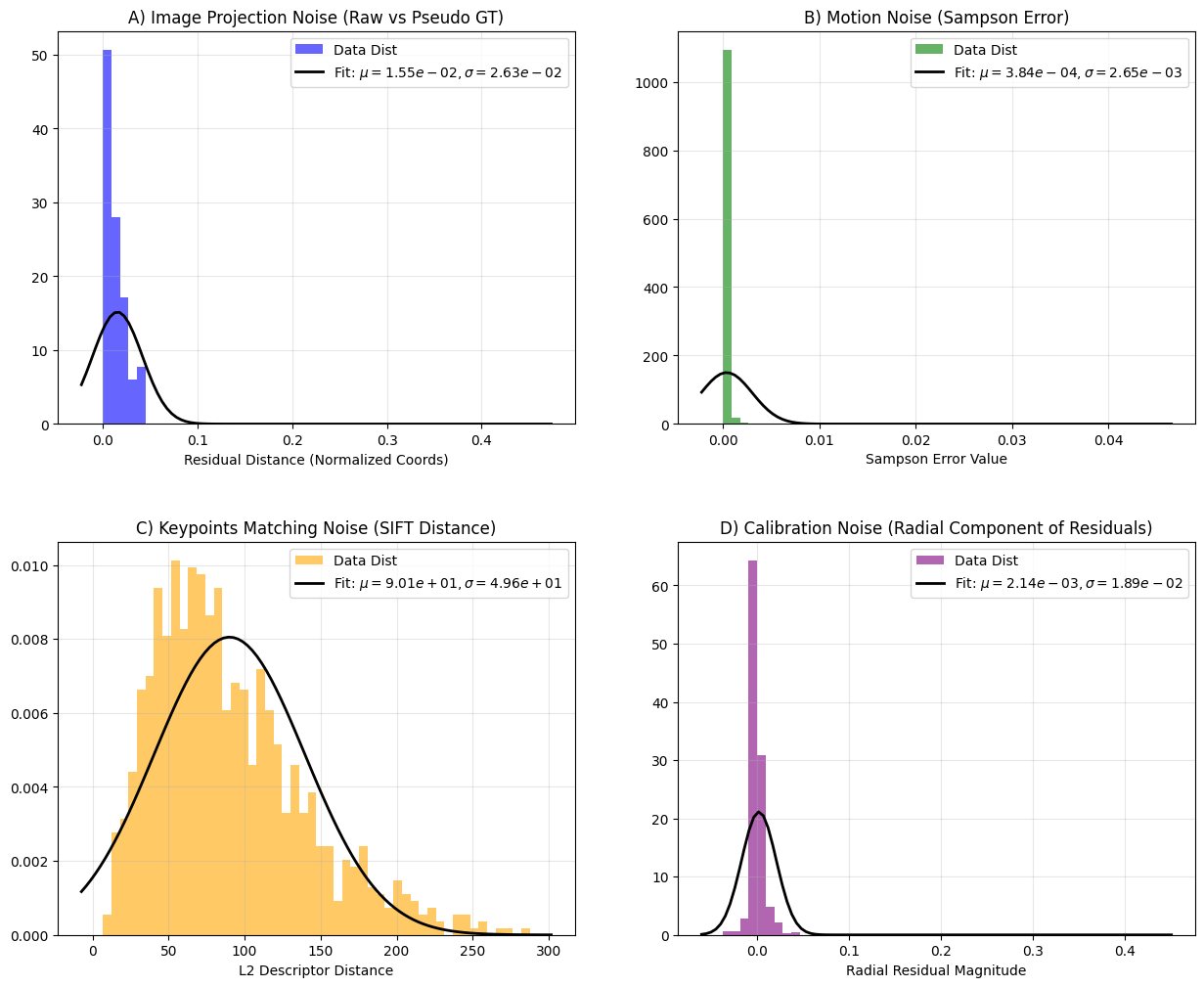}
    \caption{Noise in Images during acquisition and motion exhibiting Isotropic Gaussian Noise.}
    \label{fig:img_noise_proof}
\end{figure}

\noindent \paragraph{Projection Noise}
Image measurements are affected by sensor noise, localization uncertainty, image formation alignments and projective geometry, typically modeled as isotropic Gaussian.
\begin{equation}
\boldsymbol{\epsilon}^{proj} \sim \mathcal{N}(0, \sigma_p^2 \mathbf{I})
\end{equation}

\noindent \paragraph{Motion Noise}
Small perturbations in relative pose introduce nonlinear deviations in the epipolar geometry leading to incorrect measurements of estimated pose.
\begin{equation}
\mathbf{R}' = \mathbf{R}\exp([\delta \boldsymbol{\theta}]_\times), \quad
\mathbf{t}' = \mathbf{t} + \delta \mathbf{t}
\end{equation}
\begin{equation}
\mathbf{E}' = [\mathbf{t}']_\times \mathbf{R}'
\end{equation}
\begin{equation}
d_S(\mathbf{x}_1, \mathbf{x}_2) =
\frac{(\mathbf{x}_2^\top \mathbf{E} \mathbf{x}_1)^2}
{(\mathbf{E}\mathbf{x}_1)_1^2 + (\mathbf{E}\mathbf{x}_1)_2^2 + (\mathbf{E}^\top \mathbf{x}_2)_1^2 + (\mathbf{E}^\top \mathbf{x}_2)_2^2}
\end{equation}

\noindent \paragraph{Matching Noise}
Feature matching uncertainty introduces Non-uniform perturbations dependent on descriptor similarity.
\begin{equation}
\boldsymbol{\epsilon}^{match} \sim \mathcal{N}(0, \sigma_m^2(d)\mathbf{I}), \quad 
\sigma_m^2(d) = \alpha d + \beta
\end{equation}

\noindent \paragraph{Calibration Noise}
Imperfect calibration induces spatially varying distortions encoded from disparity maps in image measurements.
\begin{equation}
\mathbf{x}_{dist} = \mathbf{x} \left(1 + k_1 r^2 + k_2 r^4 \right)
\end{equation}
\begin{equation}
\boldsymbol{\epsilon}^{calib} = \mathbf{x}(k_1' r^2 + k_2' r^4) - \mathbf{x}(k_1 r^2 + k_2 r^4)
\end{equation}

\vspace{0.5em}
\noindent
The resulting perturbation $\boldsymbol{\epsilon}$ consists of both Gaussian and non-Gaussian components. We approximate this mixture using an isotropic Gaussian process, enabling a geometric DDPM formulation for epipolar alignment.

\subsection{Minimal Solvers for Pose Estimation}

Given corresponding points $(\mathbf{x}_1^i, \mathbf{x}_2^i)$, the essential matrix $\mathbf{E}$ satisfies:
\begin{equation}
\mathbf{x}_2^{i\top} \mathbf{E} \mathbf{x}_1^i = 0
\end{equation}

Stacking $N$ correspondences yields:
\begin{equation}
\mathbf{A}\mathbf{e} = 0
\end{equation}
where $\mathbf{e} \in \mathbb{R}^9$ is the vectorized form of $\mathbf{E}$.

\noindent \paragraph{8-Point Method}  
For $N \geq 8$, $\mathbf{E}$ is estimated by solving $\mathbf{A}\mathbf{e}=0$ followed by rank-2 enforcement via SVD.

\noindent \paragraph{5-Point Method}  
For $N = 5$, $\mathbf{E}$ is obtained from a constrained solution space satisfying $\det(\mathbf{E})=0$ and $\text{rank}(\mathbf{E})=2$.

\vspace{0.5em}
\noindent
The relative pose $(\mathbf{R}, \mathbf{t})$ is recovered from $\mathbf{E}$ via SVD:
\begin{equation}
\mathbf{E} = \mathbf{U} \boldsymbol{\Sigma} \mathbf{V}^\top
\end{equation}
\begin{equation}
\mathbf{R} = \mathbf{U}\mathbf{W}\mathbf{V}^\top, \quad 
\mathbf{t} = \mathbf{u}_3
\end{equation}
yielding up to four solutions, resolved using chirality constraints.

\subsection{Steiner Graph Construction via 3D Reconstruction}

Given matched keypoints $(\mathbf{x}_1^i, \mathbf{x}_2^i)$ between images $I_1$ and $I_2$, we lift correspondences to 3D using stereo disparity $d_i$:
\begin{equation}
z_i = \frac{fB}{d_i}
\end{equation}
where $f$ is the focal length and $B$ is the baseline. The corresponding 3D point in the camera frame is:
\begin{equation}
\mathbf{P}_i = z_i \mathbf{K}^{-1} \tilde{\mathbf{x}}_1^i
\end{equation}

We construct a graph $\mathcal{G} = (\mathcal{V}, \mathcal{E})$ over 3D points:
\begin{equation}
\mathcal{V} = \{ \mathbf{P}_i \in \mathbb{R}^3 \}
\end{equation}

Global connectivity is ensured via a Minimum Spanning Tree (MST) over Euclidean distances:
\begin{equation}
\mathcal{E}_{mst} = \arg\min_{\mathcal{T}} \sum_{(i,j)\in\mathcal{T}} \|\mathbf{P}_i - \mathbf{P}_j\|_2
\end{equation}
which serves as a Steiner tree approximation.

To preserve local geometric structure, $k$-nearest neighbor edges are added:
\begin{equation}
\mathcal{E}_{knn} = \{(i,j)\mid j \in \mathcal{N}_k(i)\}
\end{equation}

The final graph is:
\begin{equation}
\mathcal{E} = \mathcal{E}_{mst} \cup \mathcal{E}_{knn}
\end{equation}

This graph captures both global structure and local consistency for downstream geometric aggregation.

\subsection{Pseudo Ground Truth Construction for Keypoints}

Given the estimated matched keypoints $\mathbf{x}_0$ and $\mathbf{x}_1$, along with the ground truth relative pose between two frames, the pseudo ground truth correspondences are constructed using epipolar geometry constraints.

The Essential matrix is computed from the ground truth rotation $\mathbf{R}$ and translation $\mathbf{t}$ as

\begin{equation}
\mathbf{E} = [\mathbf{t}]_{\times}\mathbf{R}
\end{equation}

Using the camera intrinsic matrix $\mathbf{K}$, the Fundamental matrix is obtained as

\begin{equation}
\mathbf{F} = \mathbf{K}^{-T}\mathbf{E}\mathbf{K}^{-1}
\end{equation}

For a keypoint $\mathbf{x}_0$ in the first image, the corresponding epipolar line in the second image is computed as

\begin{equation}
\mathbf{l}_1 = \mathbf{F}\mathbf{x}_0
\end{equation}

Similarly, for a keypoint $\mathbf{x}_1$ in the second image, the corresponding epipolar line in the first image is

\begin{equation}
\mathbf{l}_0 = \mathbf{F}^{T}\mathbf{x}_1
\end{equation}

The pseudo ground truth keypoints are obtained by projecting the estimated keypoints onto their corresponding epipolar lines:

\begin{equation}
\mathbf{x}_1^{gt}
=
\mathbf{x}_1
-
\frac{\mathbf{l}_1^{T}\mathbf{x}_1}
{l_{1x}^{2} + l_{1y}^{2}}
\begin{bmatrix}
l_{1x} \\
l_{1y} \\
0
\end{bmatrix}
\end{equation}

\begin{equation}
\mathbf{x}_0^{gt}
=
\mathbf{x}_0
-
\frac{\mathbf{l}_0^{T}\mathbf{x}_0}
{l_{0x}^{2} + l_{0y}^{2}}
\begin{bmatrix}
l_{0x} \\
l_{0y} \\
0
\end{bmatrix}
\end{equation}

where $(l_x, l_y)$ denote the epipolar line coefficients. The corrected keypoints satisfy the epipolar constraint and serve as pseudo ground truth correspondences for supervision during training.

\section{Proposed Methodology}\label{sec:proposed_method}

Minimal solvers require geometrically consistent correspondences; however, real-world matches are corrupted by noise, leading to instability in pose estimation. We address this by refining correspondences, lifting them to 3D, and aggregating them through a structured graph for robust essential matrix estimation. The overall pipeline is shown in Fig.~\ref{fig:epi-block-pose}.

\begin{itemize}

\item \textbf{Step 1: Sparse Epipolar Matching} \\
Given input images $I_1$ and $I_2$, a sparse epipolar matcher produces corresponding keypoints $(\mathbf{k}_1, \mathbf{k}_2)$. These matches are geometrically constrained but still affected by noise and mismatches, motivating further refinement.

\item \textbf{Step 2: DDPM-based Keypoint Realignment} \\
To correct misalignment, we model perturbations using a diffusion process under an isotropic Gaussian assumption. For keypoints $\mathbf{k}$:
\begin{equation}
\mathbf{k}_t = \sqrt{\bar{\alpha}_t} \mathbf{k}_0 + \sqrt{1 - \bar{\alpha}_t} \boldsymbol{\epsilon}, \quad \boldsymbol{\epsilon} \sim \mathcal{N}(0, \mathbf{I})
\end{equation}
A denoising network predicts $\boldsymbol{\epsilon}_\theta(\mathbf{k}_t, t)$ to obtain refined correspondences $(\mathbf{k}_1', \mathbf{k}_2')$ that better satisfy epipolar geometry.

\item \textbf{Step 3: Stereo Depth and 3D Reconstruction} \\
Using stereo disparity, refined keypoints are lifted to 3D points $\mathbf{P}_i \in \mathbb{R}^3$ in the camera coordinate frame. This enables direct modeling of geometric relationships in Euclidean space.

\item \textbf{Step 4: Steiner Graph Construction} \\
A graph $\mathcal{G}$ is constructed over the 3D points (refer Sec.~X). A Minimum Spanning Tree ensures global connectivity, while local KNN edges preserve neighborhood structure. This provides a structured representation of correspondences for aggregation.

\item \textbf{Step 5: Essential Matrix Approximation and Relative Pose} \\
Graph features are aggregated to estimate multiple essential matrix candidates:
\begin{equation}
\mathbf{E} \in \mathbb{R}^{M \times 9}, \quad M = 5
\end{equation}
Each $\mathbf{E}_i$ is reshaped to $3 \times 3$ and passed through a differentiable SVD layer (refer Sec.~X) to recover relative pose $(\mathbf{R}_{rel}, \mathbf{t}_{rel})$. Translation is obtained up to scale.

\item \textbf{Step 6: Absolute Pose Estimation} \\
Relative transformations are accumulated to obtain absolute pose. For each step $k$:
\begin{equation}
\mathbf{T}_{rel}^{(k)} =
\begin{bmatrix}
\mathbf{R}_{rel}^{(k)} & \mathbf{t}_{rel}^{(k)} \\
\mathbf{0}^\top & 1
\end{bmatrix}
\end{equation}
\begin{equation}
\mathbf{T}_{abs}^{(k)} = \mathbf{T}_{abs}^{(k-1)} \mathbf{T}_{rel}^{(k)}
\end{equation}

\end{itemize}

The resulting trajectory is evaluated using standard metrics of: a) \textbf{Relative Pose}: RRE, RTE, Sampson Loss and b) \textbf{Absolute Pose}: ATE, APE and APE-R.

\vspace{0.5cm}

\paragraph{Image Matching module} As described in Figure~\ref{fig:epi-block-pose},Sparse Epipolar Image matching module draws inspiration from LoFTR~\cite{sun2021loftr} estimating fine feature maps $\mathbf{F}_1$ and $\mathbf{F}_2$ which are propagated to estimate probability matching matrix $\mathbf{P}_{match}$.

\paragraph{Matched Keypoint Epipolar Realignment with DDPM} Top correspondences obtained from the Sparse Epipolar Matcher are refined by enforcing ideal epipolar geometric constraints under the assumption that the camera observations are corrupted by isotropic Gaussian noise, as illustrated in Figure~\ref{fig:img_noise_proof}. Ground truth poses provided in the KITTI SLAM dataset are utilized to project matched keypoints onto their corresponding epipolar lines, thereby constructing pseudo epipolar correspondences for training supervision.

During the forward diffusion process, noise is progressively injected into the correspondence offsets, while additional stochastic perturbation is introduced to improve prediction stability. The reverse diffusion process consists of: (a) noise prediction and (b) correspondence realignment through iterative denoising of the keypoints. Further architectural details are provided in Supplementary Material Section~1.

\paragraph{Uplifted keypoint Steiner Graph, Graph Transformer and Pose Estimation}
The re-aligned keypoints, along with depth estimated using the stereo SGBM module, are projected into 3D space to construct a 3D Steiner graph and a 2D keypoint proximity graph. A Graph Propagation Transformer consisting of self-attention and cross-attention layers is designed to learn geometric feature interactions across both graph domains. The learned representations are subsequently passed to a weighted Differentiable SVD module for estimating the relative rotation matrix and translation vector. Additional architectural details are provided in Supplementary Material Section~2.

\section{Results}

The Results section begins with the implementation details of the proposed architectures, including the experimental environment, GPU specifications, and training configurations. The section is further organized into: (a) Sparse Epipolar Image Matching, (b) Keypoint Realignment using the Transformer-based DDPM module, (c) Pose estimation evaluation against state-of-the-art methods and the proposed framework on the KITTI Seq-09 and TartanAir datasets, and (d) trajectory visualization of the proposed architecture on the TartanAir dataset.

\subsection{Implementation Strategy}

The proposed framework is implemented in PyTorch using the Kaggle Jupyter environment. Training was performed on NVIDIA Tesla P100 GPUs, while inference utilized NVIDIA Tesla T4 GPUs. The complete pipeline consists of three sequential stages: a) Sparse Epipolar Matching trained for 100 epochs, b) Diffusion-based Keypoint Realignment trained for 150 epochs, and c) Graph Learning with Differentiable Pose Estimation trained for 100 epochs. For all stages, the best-performing models were selected using validation loss with a saturation/convergence threshold.

\vspace{0.3em}
\noindent
To minimize inter-stage dependency bias, each stage was optimized independently using dedicated objective functions. The Sparse Epipolar Matching stage uses Sampson loss to enforce geometric consistency between correspondences:
\begin{align}
F_1 &= \mathcal{L}_{samp}
\end{align}

The Diffusion Realignment stage jointly optimizes geometric refinement and denoising consistency using:
\begin{align}
F_2 &= w_1 \mathcal{L}_{samp} + w_2 \mathcal{L}_{ddpm} + w_3 \mathcal{L}_{rec}
\end{align}

The Graph Learning and Pose Estimation stage uses a composite objective for geometric aggregation and differentiable pose recovery:
\begin{align}
F_3 &= w_1 \mathcal{L}_{align} + w_2 \mathcal{L}_{epi} + w_3 \mathcal{L}_{svd}
\end{align}

\noindent
where $\mathcal{L}_{samp}$ denotes Sampson loss, $\mathcal{L}_{ddpm}$ is the diffusion denoising objective, $\mathcal{L}_{rec}$ enforces reconstruction consistency, $\mathcal{L}_{align}$ represents geometric alignment loss, $\mathcal{L}_{epi}$ minimizes epipolar residuals, and $\mathcal{L}_{svd}$ regularizes differentiable pose estimation. The Loss functions are further described in the Supplementary Material Section 3.

\vspace{0.3em}
\noindent
Training was performed on KITTI 2012 all training sequences till SEQ08 due to presence of ground truth pose, while inference was performed on SEQ09 dataset. Additionally, a subset of the TartanAir dataset (last samples \~ 250 frames) was used to further train the differentiable SVD module for improved inference generalization. Stage 1 uses consecutive to wide baseline initialization to learn sparse correspondences under maximal to minimal correlation conditions further refined with a spatial mask for incorrect matched correspondences. Stage 2 employs pseudo ground truth generated using estimated correspondences and ground truth poses from re-projecting correspondences on epipolar lines for diffusion-based refinement. Stage 3 uses SGBM-based stereo depth estimation with hole filling to generate dense disparity maps and 3D coordinates for scaled pose recovery in KiTTi 2012 SLAM dataset whereas corrected depths were provided by Tartan Air Dataset.

\subsection{Sparse Image Matching and Keypoint Realignment}

\begin{figure}
    \centering
    \includegraphics[width=0.9\linewidth]{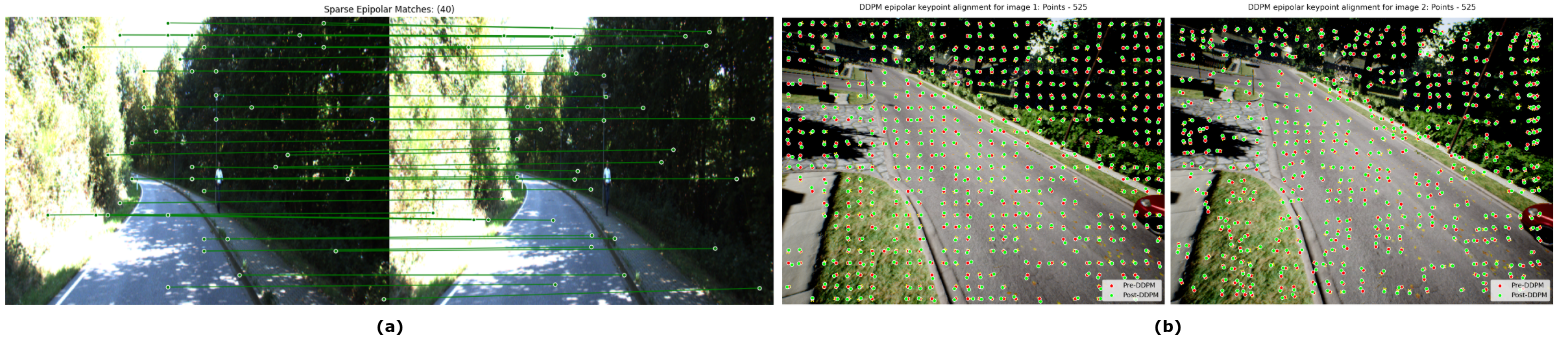}
    \caption{Result Plots: a) Sparse Epipolar Image Matching and b) Transformer DDPM keypoint Realignment}
    \label{fig:img_match_res}
\end{figure}

From the Fig~\ref{fig:epi-block-pose}, as described in section~\ref{sec:proposed_method} both Image matching and DDPM based keypoint realignment in reverse diffusion process by predicting isotropic gaussian noise aligning to epipolar geometry to refine the keypoints as shown in Fig~\ref{fig:img_match_res}.

\begin{figure}[t]
\centering
\begin{tikzpicture}
\begin{axis}[
    width=0.7\linewidth,
    height=4cm,
    thick, 
    xlabel={Frame Number},
    ylabel={Sampson Error},
    xmin=0,
    xmax=9,
    ymin=0,
    ymax=8, 
    xtick={0,1,2,3,4,5,6,7,8,9},
    ytick={0,2,4,6,8},
    grid=major,
    legend style={
        at={(0.5,-0.5)},
        anchor=north,
        legend columns=2
    }
]

\addplot[
    color=red,
    mark=*,
    line width=1pt
]
coordinates {
(0,3.718)
(1,2.088)
(2,1.922)
(3,3.257)
(4,4.552)
(5,3.462)
(6,5.422)
(7,5.632)
(8,6.189)
(9,5.883)
};

\addplot[
    color=blue,
    mark=square*,
    line width=1pt
]
coordinates {
(0,3.643)
(1,1.966)
(2,1.905)
(3,3.135)
(4,4.529)
(5,3.899)
(6,5.189)
(7,5.361)
(8,6.357)
(9,5.701)
};

\legend{Initial Sparse Matches, DDPM based Refinement}

\end{axis}
\end{tikzpicture}
\caption{Comparison of correspondence sampson error from Sparse Epipolar Matcher outputss and after DDPM-based keypoint refinement across multiple frame pairs.}
\label{fig:ddpm_comparison}
\end{figure}
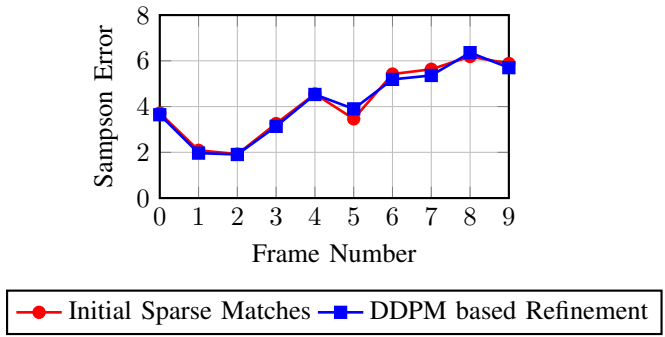


The Figure~\ref{fig:ddpm_comparison} compares Sampson Loss from Sparse Epipolar matcher supervised by epipolar geometry constraints to Sampson Loss obtained from the estimation of isotropic gaussian noise during reverse diffusion process used in refining matched keypoints. Thus from the plot DDPM module reduces Sampson Loss for matched keypoints.

\subsection{Relative and Absolute Pose Metrics Evaluation}

The proposed methodology is evaluated on the SEQ09 of KiTTi Stereo SLAM and Tartan Air Neighborhood dataset on Relative and Absolute Pose Metrics. The following results have been presented in Table~\ref{tab:kitti_seq09_results} and Table~\ref{tab:tartan_air_results_full}.

\begin{table*}[t]
\centering
\caption{Inference results on the KiTTi SEQ09 test dataset over the first 100 samples.}
\label{tab:kitti_seq09_results}
\resizebox{\textwidth}{!}{
\begin{tabular}{l c c c c c c}
\toprule
\textbf{Module Name} & \textbf{RRE ($^\circ$)} & \textbf{RTE (m)} & \textbf{Sampson Loss} & \textbf{ATE (m)} & \textbf{APE (m)} & \textbf{APE-R ($^\circ$)} \\
\midrule
Sparse Epi Matcher + RANSAC & 16.2001 & 0.2943 & 123.43759 & \textbf{26.0671} & \textbf{32.9066} & 99.5288 \\
Sparse Epi Matcher + MAGSAC & 19.3480 & 0.3181 & 146.18692 & 36.7278 & 46.2647 & 115.5531 \\
Sparse Epi Matcher + MAGSAC++ & 18.6822 & 0.3187 & 138.46090 & 33.7775 & 43.7449 & 117.5174 \\
Sparse Epi Matcher + DSAC & 17.2418 & 0.3218 & 139.83944 & 29.0708 & 35.4699 & 104.3801 \\
\midrule
Sparse Epi Matcher + DDPM + RANSAC & 8.9481  & 0.2906 & 139.83966 & 28.2286 & 33.5058 & 98.3069 \\
Sparse Epi Matcher + DDPM + MAGSAC & 6.9297  & 0.2935 & 132.46495 & 36.8140 & 48.4406 & 125.7633 \\
Sparse Epi Matcher + DDPM + MAGSAC++ & 8.8009  & 0.3093 & 140.45212 & 46.2614 & 58.4247 & 119.5690 \\
Sparse Epi Matcher + DDPM + DSAC & 8.1603  & 0.2875 & 134.48266 & 33.8399 & 42.7565 & 129.1885 \\
\midrule
Deep Fundamental Matrix & \textbf{1.0724}  & 0.6937 & 140.8431 & 40.5094 & 46.7136 & \textbf{84.3638} \\
Sparse Epi Matcher + DDPM + GraphDiffSVD & 8.6926  & \textbf{0.2568} & \textbf{122.9859} & 27.2388 & 33.9787 & 116.5776 \\
\bottomrule
\end{tabular}
}
\end{table*}

\begin{table*}[t]
\centering
\caption{Tartan Air Neighbourhood Inference over 100 samples}
\label{tab:tartan_air_results_full}
\resizebox{\textwidth}{!}{
\begin{tabular}{l c c c c c c}
\toprule
\textbf{Module Name} & \textbf{RRE ($^\circ$)} & \textbf{RTE (m)} & \textbf{Sampson Loss} & \textbf{ATE (m)} & \textbf{APE (m)} & \textbf{APE-R ($^\circ$)} \\
\midrule
Sparse Epi Matcher + RANSAC & 5.0621 & 0.1493 & 28.45168 & 7.4902 & 7.9161 & 122.7500 \\
Sparse Epi Matcher + MAGSAC & 4.9927 & 0.1518 & 29.06684 & 7.7595 & 8.1806 & 127.8952 \\
Sparse Epi Matcher + MAGSAC++ & 5.0918 & 0.1465 & 35.71909 & 6.8685 & 7.2854 & 123.5199 \\
Sparse Epi Matcher + DSAC & 4.4116 & 0.1511 & \textbf{27.64794} & 7.9721 & 8.4190 & 122.9739 \\
\midrule
Sparse Epi Matcher + DDPM + RANSAC & 5.2032 & 0.1465 & 33.82532 & 7.7132 & 8.2823 & 98.6421 \\
Sparse Epi Matcher + DDPM + MAGSAC & 5.3364 & 0.1467 & 27.95476 & 7.4155 & 7.8532 & 118.8436 \\
Sparse Epi Matcher + DDPM + MAGSAC++ & 5.1259 & 0.1478 & 29.37567 & 7.7961 & 8.2467 & 115.2226 \\
Sparse Epi Matcher + DDPM + DSAC & 5.0943 & \textbf{0.1457} & 29.85782 & 7.1940 & 7.6359 & 126.5704 \\
\midrule
Sparse Epi Matcher + DDPM + GraphDiffSVD & \textbf{2.5485} & 0.2144 & 33.23149 & \textbf{5.5960} & \textbf{6.0302} & \textbf{79.6835} \\
\bottomrule
\end{tabular}
}
\end{table*}

The inference metrics indicate that the estimated Essential matrix serves as the fundamental basis for both Absolute Pose estimation and Sampson error computation. From the KITTI evaluation results presented in Table~\ref{tab:kitti_seq09_results}, it is observed that the Sparse Epipolar Matcher combined with RANSAC achieves superior performance for absolute pose estimation. In contrast, for the TartanAir dataset, despite exhibiting comparatively higher Sampson error values, the proposed approach demonstrates improved robustness in estimating the initial absolute pose, as shown in Table~\ref{tab:tartan_air_results_full}.

\begin{figure*}
    \centering
    \includegraphics[width=0.75\linewidth]{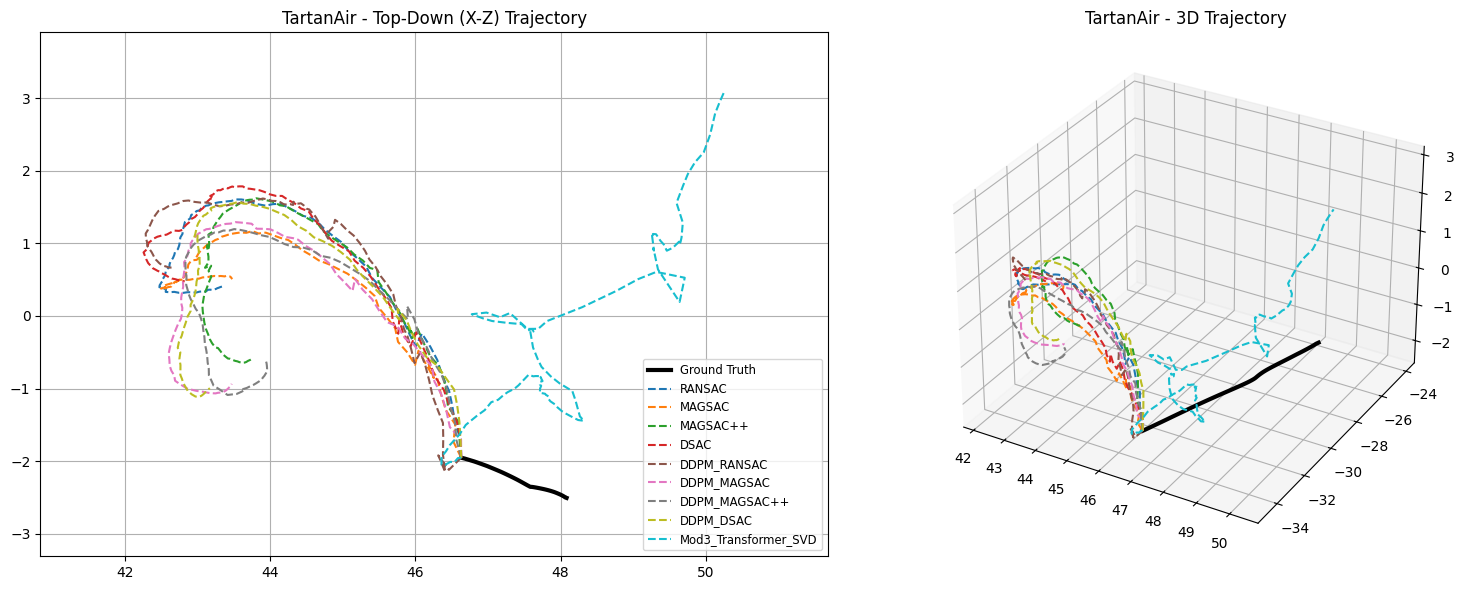}
    \caption{Pose Estimation Evaluation (Absolute Pose) over defined methods in Table~\ref{tab:tartan_air_results_full} : a) X-Z 2D trajectory of the models and b) 3D Cartesian Trajectory Plot}
    \label{fig:tartan_air_plot}
\end{figure*}

Since the proposed framework achieved the best overall performance on the TartanAir dataset, the corresponding trajectory visualization is presented in Figure~\ref{fig:tartan_air_plot}.

\section{Conclusion}

In this work, we presented a geometry-driven relative pose estimation framework combining Sparse Epipolar Matching, Transformer-based DDPM keypoint refinement, and a Graph Propagation Transformer integrated with a differentiable SVD solver. The proposed framework focuses on learning compact and geometrically consistent correspondences instead of relying on dense matching or direct pose regression. Experimental evaluation on the KITTI and TartanAir datasets demonstrates that the proposed approach achieves the lowest Sampson error on the KITTI dataset while providing the most stable absolute pose estimation on the TartanAir dataset. The results further show that diffusion-based correspondence refinement and graph-based subset selection improve geometric consistency and reduce correspondence redundancy under challenging viewpoint and motion variations.

\section{Future Work}

Future work will focus on extending the proposed framework toward real-time Visual SLAM and Visual Odometry systems operating under dynamic and degraded environments. Additional directions include integrating event-based cameras and IMU fusion for robust motion estimation in high-speed scenarios, improving scale consistency through temporal graph propagation, and exploring lightweight graph transformer architectures for deployment on embedded robotic and autonomous navigation platforms.

\nocite{*}

\bibliographystyle{IEEEtran}
\bibliography{ref} 
\vfill

\end{document}